%% file: main.tex
\documentclass{article}

\usepackage{microtype}
\usepackage{graphicx}
\usepackage{subfigure}
\usepackage{enumitem}
\usepackage{booktabs} 
\usepackage{multirow}
\usepackage{hyperref}
\usepackage{algorithmic}
\usepackage{float} 

\usepackage[accepted]{icml2024}

\usepackage{amsmath}
\usepackage{amssymb}
\usepackage{mathtools}
\usepackage{amsthm}

\usepackage[capitalize,noabbrev]{cleveref}

\theoremstyle{plain}

\theoremstyle{definition}

\theoremstyle{remark}

\newcommand{\ifcommentsenabled}[1]{}

\definecolor{todo_color}{rgb}{1.0,0,0.0}
\definecolor{help_color}{rgb}{0.7,0.6,0.2}

\definecolor{edited_color}{rgb}{.5,.7,.1}
\definecolor{qihang_color}{rgb}{.6,.4,.05}
\definecolor{chengcheng_color}{rgb}{0.35,0,0.75}
\definecolor{bugra_color}{rgb}{0,0.35,0}
\definecolor{shugao_color}{rgb}{0,0.35,0.35}
\definecolor{yanchao_color}{rgb}{0,0,0.85}

\newcommand{\chengcheng}[1]{\ifcommentsenabled{\textcolor{chengcheng_color}{C: #1}}}

\usepackage[textsize=tiny]{todonotes}

\icmltitlerunning{
BID: Boundary-Interior Decoding for Unsupervised Temporal Action Localization Pre-Training}

\begin{document}

\twocolumn[

\icmltitle{
BID: Boundary-Interior Decoding \\
for Unsupervised Temporal Action Localization Pre-Training}



\icmlsetsymbol{equal}{*}

\begin{icmlauthorlist}
\icmlauthor{Qihang Fang}{hku}
\icmlauthor{Chengcheng Tang}{rl}
\icmlauthor{Shugao Ma}{rl}
\icmlauthor{Yanchao Yang}{hku}

\icmlcorrespondingauthor{Chengcheng Tang}{chengcheng.tang@meta.com}
\icmlcorrespondingauthor{Yanchao Yang}{yanchaoy@hku.hk}

\end{icmlauthorlist}

\icmlaffiliation{hku}{The University of Hong Kong}
\icmlaffiliation{rl}{Meta Reality Labs}
%

\icmlkeywords{Machine Learning, ICML}

\vskip 0.3in
]



\printAffiliationsAndNotice{}  


\input{sections/00_abstract}
\input{sections/01_introduction}
\input{sections/02_related_work}

\input{sections/03_method}
\input{sections/04_exp}

\input{sections/05_conclusion}

\bibliography{main}
\bibliographystyle{icml2024}

\input{sections/06_appendix}

\end{document}


%% file: sections/00_abstract.tex
\begin{abstract}
Skeleton-based motion representations are robust for action localization and understanding for their invariance to perspective, lighting, and occlusion, compared with images.
Yet, they are often ambiguous and incomplete when taken out of context, even for human annotators.
As infants discern gestures before associating them with words, actions can be conceptualized before being grounded with labels.
Therefore, we propose the first unsupervised pre-training framework, {\it Boundary-Interior Decoding (BID)}, that partitions a skeleton-based motion sequence into discovered semantically meaningful {\it pre-action} segments.
By fine-tuning our pre-training network with a small number of annotated data, we show results out-performing SOTA methods by a large margin.

\end{abstract}

%% file: sections/01_introduction.tex
\section{Introduction}
Developing a human-centric AI assistant providing timely assistance and feedback, as a collaborator for constructing a project or an instructor for learning a skill, requires a system that thoroughly understands human actions. A premise to action understanding is the ability to recognize and segment human actions from a long sequence, through the temporal action localization (TAL) task, which is critical as a standalone application and foundational for downstream tasks such as alignment and detailed action recognition.

\label{sec:intro}
\begin{figure}[htb]
    \centering
    \includegraphics[width=.95\linewidth]{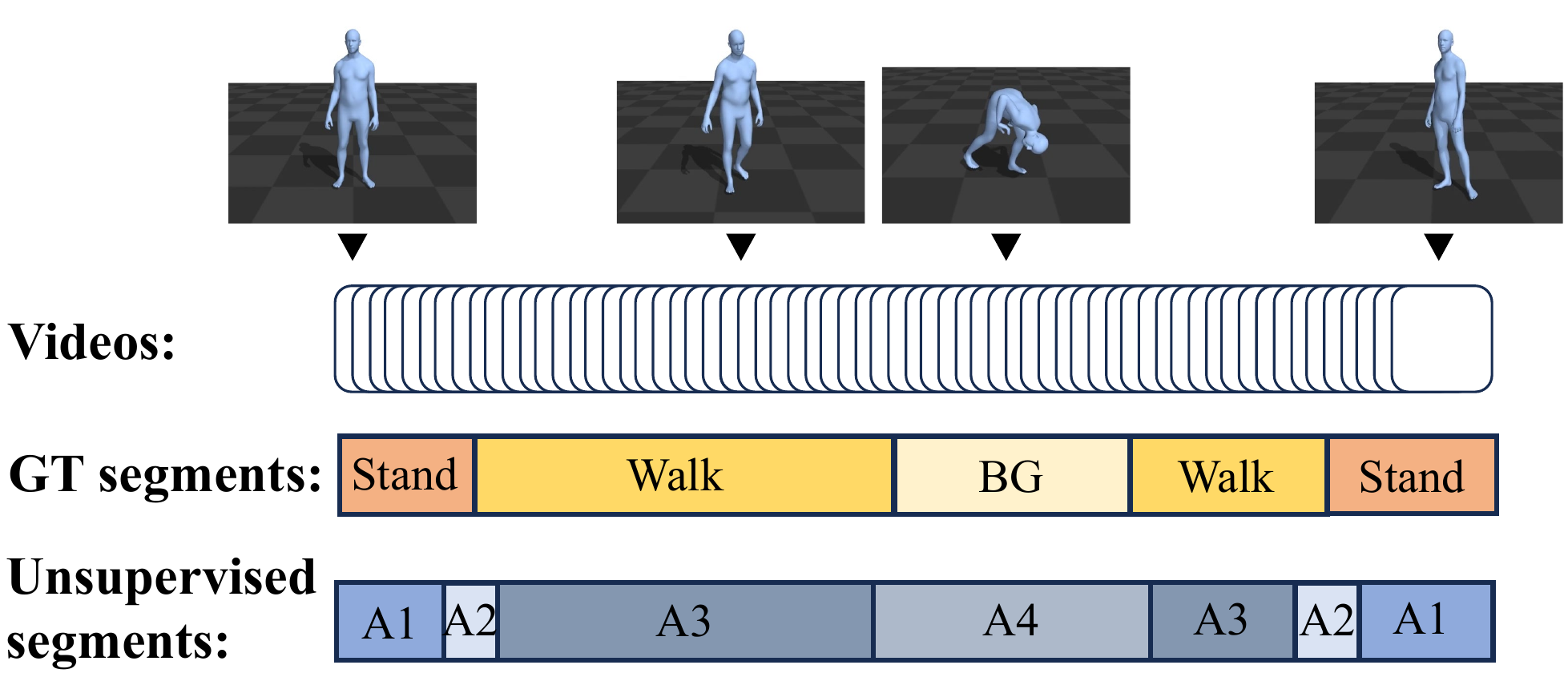}
    \caption{
We propose an unsupervised method to partition a skeletal motion to semantically meaningful action segments, which we name as {\it pre-action} segments and classes to differentiate from those defined by human annotators.
We study how the discovery of pre-actions can help improve label efficiency with fine-tuning on limited human labels, for temporal action localization.
    }
    \label{fig:teaser}
\end{figure}

Model pre-training, which involves learning representations from large-scale data, has become a common practice for vision-based temporal action classification. This approach has proven to be effective for temporal action classification (TAC) \cite{PAL-10, PAL-55}, which aims to predict video-level or instance-level action labels for trimmed data. However, TAC disregards the start time and duration of each action, which are crucial for temporal action localization (TAL).
As a result, the pre-training features derived from TAC are insufficient for TAL, which needs to infer the start time and duration for each action with untrimmed sequences.
In recent years, some researchers \cite{PAL-1,PAL-61,PAL-62} begin to address this issue by developing specific methods for TAL tasks. However, these methods heavily rely on large-scale annotated video data, which is challenging and expensive to annotate accurately with untrimmed videos, particularly when facing a large range of ambiguous action types.
\cite{zhang2022unsupervised} proposed the first unsupervised pre-training method for vision-based TAL. Their approach involves designing a contrast learning task disentangling interleaved videos from different sequences through data augmentation. However, inserting and rediscovering artificially cut clips from one video into another without semantics disrupts the continuity of the action sequence. As a result, this approach faces challenges in locating accurate segmentation boundaries. 

While vision-based temporal action localization (TAL) has made significant progress, it has limitations in terms of robustness towards viewpoints, lighting conditions, and image noise. In contrast, 3D skeleton-based data provides a more reliable representation for applications such as animation, augmented reality, and gaming. As a result, there is a growing interest in action understanding for skeleton-based data \cite{punnakkal2021babel,yu2023frame}. 
At the same time, skeleton-based tasks require a specific focus on human posture during actions due to the absence of visual context. For instance, vision-based tasks can determine if an athlete is playing basketball based on the presence of a basketball court, whereas such contextual information is not available for skeleton-based tasks. Due to the unique challenges, most existing skeleton-based methods \cite{punnakkal2021babel,vemulapalli2014human,du2015hierarchical} primarily focus on TAC instead of TAL. Recently, \cite{yu2023frame} proposed the first skeleton-based TAL algorithm that incorporates a weakly supervised pre-training step using sequence-level annotation available to the input untrimmed videos. However, as the sequence-level annotation often lacks granularity and temporal resolution, the pre-training results are susceptible to long-tail effects, facing challenges in capturing shorter and infrequent actions.

To address the issues mentioned above, we propose {\it BID} which is an unsupervised pre-training method for skeleton-based temporal action localization. We regard human actions as different flow fields and train a decoding function to represent these flow fields by discrete latent codes. Through these flow fields, we can discover meaningful action segmentation and classification without human annotation. To achieve this, we enforce that the discrete latent codes inform the states
within each action segment, for which we train an {\it Interior Decoder} that can inpaint the random masked sequence based on the latent codes. Besides, we also require the discrete latent codes to inform or help select the end state of the action, for which we instantiate a {\it Boundary Decoder} to predict the end state for each action segment. By supervising these two decoders, we can achieve semantically meaningful segmentation and classification. Once we obtain the discrete action representations of human motion, we can employ them for Temporal Action Localization (TAL) with fine-tuning on a limited number of annotations


To evaluate the effectiveness of our proposed method for TAL, we conduct experiments using the BABEL dataset \cite{punnakkal2021babel} and compared it with another skeleton-based method \cite{yu2023frame} and other vision-based methods \cite{zhang2022unsupervised, zhang2021cola, huang2021foreground, rajasegaran2023benefits} adapted to the skeleton-based settings.
We show that our method outperforms existing methods, indicating the semantic meaningfulness and groundability of pre-action segments and classes emerged through our proposed method of unsupervised skeleton-based temporal action localization.

We summarize the contributions of this paper as follows:
\begin{itemize}[leftmargin=*]
    \setlength{\itemsep}{0pt}
    \setlength{\parsep}{0pt}
    \setlength{\parskip}{0pt}

    \item We introduce {\it BID}, which, to our best knowledge, the first unsupervised skeleton-based method for temporal action localization through a pre-training framework.
    \item By modeling actions as bounded and directed flow fields, we propose two optimization targets for the pre-training framework, focusing on the interior and boundary of segments respectively.
    \item We conduct comprehensive evaluations and comparisons of our approach, outperforming SOTA methods by 20\%. We study emerged pre-action classes and segments qualitatively, quantitatively, and visually, validating that, without any human annotation, the proposed pre-training framework develops a surprisingly strong sense of semantic understanding that correlates with human annotated action class labels and segments.
\end{itemize}

%% file: sections/02_related_work.tex
\section{Related Work}
\label{sec:related}

\paragraph{Unsupervised Representation Learning.}
In the recent era, computer vision rediscovered the power of unsupervised learning ~\cite{caron2020unsupervised, chen2020simple, chen2020exploring, chen2020improved, bootstrap_latent, he2020momentum, Wu_2018_CVPR}. 

In ~\cite{qian2021spatiotemporal}, the authors propose a novel framework that leverages the temporal dynamics and spatial features in videos to learn rich, discriminative representations. In ~\cite{wang2020self},  the authors introduce a self-supervised learning approach for video representation that is based on predicting the pace of video sequences, training a model to distinguish between normal and temporally altered (e.g., sped-up or slowed down) video clips. In ~\cite{yang2020video},  the authors propose a technique that aligns the representations of video segments with varying tempos, ensuring that the learned representations are consistent across different speeds of motion. In different contexts, contrastive learning with videos has shown promise for a wide range of vision-based tasks. For example, employing positive-augmented contrastive learning to better distinguish between closely related captions and visual content enhances image and video captioning evaluation~\cite{Sarto_2023_CVPR}; a semi-supervised contrastive learning framework for ego-vehicle action recognition also improves the accuracy and robustness of action recognition in autonomous driving scenarios~\cite{Noguchi_2023_WACV}.

\paragraph{Temporal Action Classification.}
Temporal action classification (TAC) is also known as action recognition, which aims to predict sequence-level or video-level action labels for trimmed data.
Earlier works on action recognition rely on handcrafted features \cite{vemulapalli2014human, liu2016spatio, dollar2005behavior, wang2013action}. Among the tasks of video and action understanding, TAC of trimmed data or vision-based action recognition \cite{shi2019two, PAL-21, rajasegaran2023benefits} is a popular task. At first, 3D convolution networks are the commonly adapted backbone for vision-based understanding \cite{taylor2010convolutional, tran2015learning}. Considering that 3D convolution networks treat the spatial and temporal information similarly, optical flow \cite{shi2019two} is introduced in this field. SlowFast \cite{PAL-21} utilizes video streams but at different frame rates to learn to fuse spatial and temporal information. Besides, Another line of works \cite{pan2021actor, wang2018non, wang2018videos} focuses on the relationship between actors and objects rather than the model structure.

For skeleton-based action recognition, some work \cite{du2015hierarchical, liu2016spatio} utilizes the RNN-based method to extract the temporal information from the human joint sequences. CNN-based methods \cite{liu2017enhanced,ke2017new} are also introduced in this task. Furthermore, ST-GCN \cite{yan2018spatial} proposes to model the skeleton-based data to graph structure which can be cooperated with ST-GCN, which utilizes the connectivity of human joints. Adaptive graph convolutional network (AGCN) \cite{shi2019two} extends ST-GCN by parameterizing the graph structure of skeleton data and embedding it into the network. Recently, unsupervised methods for action recognition based on ST-GCN have been developed, such as AimCLR \cite{guo2022contrastive} and ActCLR \cite{lin2023actionlet}. These methods aim to learn representations without relying on annotated labels.

\paragraph{Temporal Action Localization (TAL) Task.}
Unlike action recognition task \cite{PAL-10,PAL-21,PAL-35,PAL-49,PAL-55}, the target of TAL is to temporally localize the action of interest in untrimmed videos. In general, TAL tasks can be split into two different sub-tasks, temporal action segmentation and temporal action classification. Some previous methods utilize temporal anchor to compose action window \cite{PAL-6, PAL-30, PAL-37}. Some other works segment the actions by directly predicting the boundary probabilities \cite{PAL-36}. Furthermore, the combination of segmentation and classification is also explored \cite{PAL-36,xu2017r,xu2020g}.

Due to the dataset scale and GPU memory constraints, some works \cite{PAL-1,PAL-61, PAL-62} tackle pre-train models on large-scale trimmed TAC datasets and then use it to extract frame-level or segment-level features in untrimmed TAL data. Motivated by the recent success of unsupervised learning, the approach by~\cite{zhang2022unsupervised} introduces a novel method for segmenting actions using unsupervised learning. They introduce a novel self-supervised pretext task called Pseudo Action Localization (PAL). 
Subsequently, in ~\cite{yu2023frame}, the authors proposed a novel problem of skeleton-based weakly-supervised temporal action localization (S-WTAL), recognizing and localizing human action segments in untrimmed skeleton videos given only the video-level labels.


%% file: sections/03_method.tex
\begin{figure*}[htb]
    \centering
    \includegraphics[width=1\linewidth]{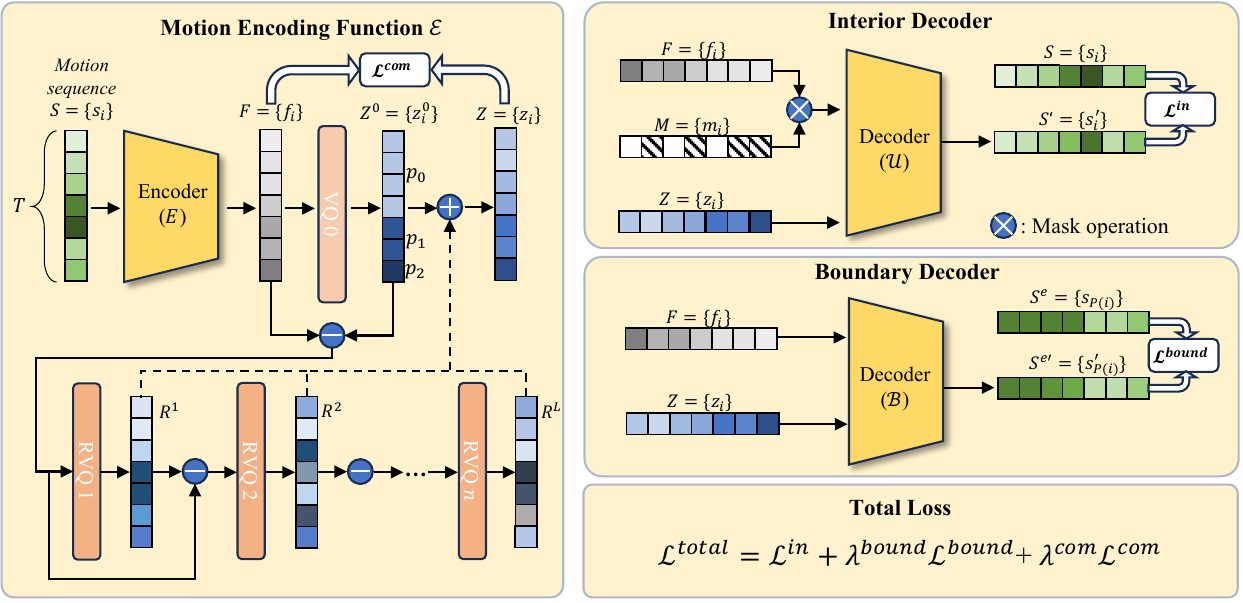}
    \vspace{-5mm}
    \caption{
    An overview of the proposed framework, 
    which consists of a motion encoding function $\mathcal{E}$, an interior decoder $\mathcal{U}$, and a boundary decoder $\mathcal{B}$. The motion encoding function $\mathcal{E}$ is composed of a linear encoder $E$ and a residual VQ (RVQ) module, which encodes the input motion sequence $S$ into discrete latent codes. The frames with the same discrete latent code predicted by the VQ layer are considered with the same action class. The summation of the discrete latent codes from the VQ layer and the RVQ layers is used as the class representation of the motion frames. Besides the commitment loss for training the RVQ module, we devise two optimization objectives that encourage the class representation to be informative about the state transition within an action and the ending state of that action.   
    }
    \label{fig:pipeline}
\end{figure*}


\section{Problem Statement}
\label{sec:statement}
\subsection{Skeleton-Based Temporal Action Localization}

Give a skeleton-based untrimmed motion sequence $S=\{s_i\}_{i=1}^T \in \mathbb{R}^{T \times J}$, where $T$ is the number of frames for the temporal sequence and $J$ is the total number of degrees of freedom of skeletal joints,
the goal of the temporal action localization (TAL) task is to divide the sequence to mutually exclusive segments with action labels, $\{t^b_m, t^e_m, a_m\}_{m=1}^M$, where $t^b_m, t^e_m, a_m$ are the beginning, end timestamps and action class of the $m$-th predicted action.

Traditionally, the action labels are predefined and manually labeled in a dataset. 
In this work, we study a new problem where the action labels are not available, 
and we propose a unsupervised method to discover semantically meaningful actions (or segments of a motion trajectory), which we name as {\it pre-action} segments and classes to differentiate from those defined by human annotators.
We will then study how the discovery of pre-actions can help improve label efficiency with fine-tuning on limited human labels.

\subsection{Action Flow Field}
From the Lagrangian point of view, 
the motion of an object moving in space can be represented as a flow field, i.e., a function that describes the relationship between time and the object's position:
\begin{equation}
    x_t = \mathcal{G}(t), \label{eq:flowfield}
\end{equation}
where $\mathcal{G}(\cdot)$ is the flow that characterizes the position of the object $x_t$ at time $t$ in a reference coordinate frame.

Human motion sequences can be similarly represented as a flow field of human bodies moving through space and time \citep{rajasegaran2023benefits}, 
with different actions represented by distinctive flow fields. 
Thus, we utilize the concept of flow fields in Equation~\ref{eq:flowfield} to represent human actions, and reformulate it in the context of skeleton-based TAL:
\begin{equation}
    s_t = \mathcal{G}^{a_m}(t) ,  t^b_m \leq t \leq t^e_m,
\end{equation}
where $\mathcal{G}^{a_m}$ is the flow field characterizing action $a_m$ 
and $s_t$ is the motion frame at the timestamp $t$. 
We call the flow field representing human actions the {\it action flow field}.

\section{Unsupervised TAL Pre-Training}
We aim to build an unsupervised skeleton-based pre-training framework that divides an input motion sequence into semantically meaningful segments with action features categorizing action flow fields without annotation.

Specifically, 
we train a neural motion encoding function, $\mathcal{E}$, unsupervised, 
to partition the input sequence $S$ to multiple segments $\{p_m\}^M_{m=1}$ with $p_m=\{s_t\}, t=t_m^b...t_m^e$, whose action label is $a_m$.
The number of action classes $N$ is predefined and equal to the maximum possible number of action classes in the dataset. 
Due to the lack of ground-truth labels, 
these segments and actions may not be completely aligned with human annotations. 
And as mentioned above, we call these segments, $\{p_m\}^M_{m=1}$, {\it pre-action segments} or actions, $\{a_m\}^M_{m=1}$, {\it pre-action classes}.


To enable training,
we propose that the {\it pre-action class} ${a_m}$
should help instantiate a flow field function $\mathcal{G}^{a_m}$ that can be used to reconstruct or predict the temporal movements of the human body within this action clip. 
Through learning for these flow field functions, we can discover meaningful {\it pre-action segments} and {\it pre-action classes}. 
More explicitly,
we enforce that the pre-action class should inform the states within each action clip as much as possible, for which we train an {\it Interior Decoder} that reconstructs the state transitions within the segment from the masked clip and the discovered class representation.
Moreover, 
the discovered action class should inform or help select the end state of the action, for which we instantiate a {\it Boundary Decoder} that predicts the terminating state $s_{t_m^e}$ from the current state and the class.
An overview of the training scheme is shown in Figure~\ref{fig:pipeline}.
Once we obtain the discrete action representations of human motion, we can employ them for Temporal Action Localization (TAL) with fine-tuning on a limited number of annotations.
Next, we detail the neural architecture for learning the encoding function $\mathcal{E}$.

\subsection{Motion Encoding Function}
Vector Quantized Variational AutoEncoder (VQ-VAE) \citep{van2017neural} are widely used to represent human motion as discrete latent codes \citep{van2017neural,zhang2023t2m,jiang2023motiongpt} in the literature of motion generation and synthesis. 
We also leverage VQ-VAE to construct the temporal-spatial encoding function $\mathcal{E}$ for extracting discrete latent codes to represent different actions. 
The VQ-VAE-based encoding function $\mathcal{E}$ is composed of a linear encoder $E$ and a VQ layer. 
The VQ layer contains a codebook $c$ to quantize the input continuous features into discrete latent codes. 
Specifically, $E$ encodes the input sequence $S$ into a sequence of features $F \in \mathbb{R}^{T \times D}$, with $T$ per-frame features $\{f_i\}_{i=1}^T$, where $f_i \in \mathbb{R}^D$:
\begin{equation}
\label{eq:encoding}
F = E(S), \text { where } F \in \mathbb{R}^{T \times D}.
\end{equation}
The codebook $c \in \mathbb{R}^{K \times D}$ represents different action flow fields, where $K$ is a predefined number of different discrete codes or motion modes and $c_k \in \mathbb{R}^D$ is the k-th action representation. 
The VQ layer encodes the sequence feature $F$ to a discrete latent code(s) $Z  \in  \mathbb{R}^{T \times D}$ through codebook $c$ by projecting each per-frame feature $f_i$ to its nearest code:
\begin{align}
VQ\left(f_i\right)=c_k, \text { where } k=\operatorname{argmin}_j\left\|f_i-c_j\right\|_2^2,
\end{align}
where $VQ$ represents the VQ layer, $Z$ is the ensemble $\{VQ(f_i)\}_{i=1}^{T}$, and $k$ is the code index corresponding to a {\it pre-action class}. We consider adjacent motion frames assigned with the same {\it pre-action class} constituting a {\it pre-action segment}.
To facilitate learning and guarantee convergence, we impose a commitment loss following \citep{van2017neural}:
\begin{equation}   
\label{eq:com}
\mathcal{L}^{com}=\sum\limits_{i=1}^{T}\left\|f_i - sg(VQ(f_i))\right\|^2_2,
\end{equation}
where $sg$ is the stop-gradient operation. 




\subsection{Residual VQ-VAE}

In practice, we find two problems with the VQ-VAE-based encoding: 1) Codebook collapse, where only a small number of codes in the codebook are activated, and a large number of codes are rarely used; 
2) Imbalance of high-frequency information and action semantics. 
To retain enough high-frequency information for good reconstruction quality, one could increase the number of codes in the codebook to avoid over-compress. However, increasing the number of discrete latent codes will make the {\it pre-action segments} become over-fragmented and semantically ambiguous. 

Similar phenomena are also reported in motion generation \citep{yao2023moconvq}, audio compression \citep{zeghidour2021soundstream}, and image generation \citep{lee2022autoregressive}. 
To solve these problems, Residual VQ-VAE \cite{lee2022autoregressive} has been proposed, which employs a residual architecture with multiple quantization layers to enhance the utilization of the codebook and stabilize the training process. 
In addition to the standard VQ-VAE, 
we also employ a series of residual vector quantization (RVQ) layers to resolve the above issues (Figure~\ref{fig:pipeline}). 
More explicitly, 
we first utilize the VQ layer to quantize input feature $F$ to discrete latent code $Z^0$. 
Then, an RVQ layer is utilized to quantize the residual $F - Z^0$ to discrete latent code $R^1$. Subsequent, the $l$-th RVQ layer is utilized to quantize the residual $F - Z^0 - \sum\limits_{i=1}^{l-1}R^i$ to discrete latent code $R^l$. 
With the RVQ layers, the final quantized code $Z$ for input $F$ is the sum of all the discrete code,  
\begin{equation}
\label{eq:sumz}
    Z = Z^0 + \sum\limits^{L}_{l=1}{R^l},
\end{equation}
where $L$ is the number of RVQ layers. 

In \citep{lee2022autoregressive}, 
the VQ layer and all the RVQ layers share the same codebook to increase the usage of the codebook. However, repeatedly extracting discrete codes from the same codebook makes it difficult to define the {\it pre-action classes}. Considering that the Residual VQ-VAE builds a coarse-to-fine representation of the entire latent space \citep{yao2023moconvq,lee2022autoregressive}, we initialize two different codebooks $c^{VQ}$ and $c^{RVQ}$ for the VQ layer and RVQ layers respectively. The first codebook is only used for the VQ layer, while the frames with the same discrete code quantized by the VQ layer are considered with the same {\it pre-action class}. The second codebook is shared by all the RVQ layers for retaining details important for reconstruction but not high-level semantics.

\subsection{Boundary and Interior Decoding for Pre-Actions}
\label{sec:flow}
To partition an input motion sequence 
to semantically meaningful {\it pre-action} segments and classes, we optimize the discrete latent codes representing the action flow fields for the informativeness of both the state transitions within an action (Interior Decoding) and the ending state of that action (Boundary Decoding).

\textbf{Interior Decoding}. 
Given an untrimmed motion sequence $S$ with several {\it pre-action segments} $p_m=\{t^b_m, t^e_m, a_m\}$ (abuse notation for simplicity), 
the motion frames $\{s_i\}_{i=t^b_m}^{t^e_m}$ belonging to a pre-action segment $p_m$ can be regarded as state transitions in the interior of the segment governed by the action flow field defined by $a_m$. 
Thus, we propose that the action flow field generated from the discrete latent codes $\{z_i\}_{i=t^b_m}^{t^e_m}$ (action representation) should be as close as possible to the input states $\{s_i\}_{i=t^b_m}^{t^e_m}$. 

A naive approach 
to achieving this optimization goal is to 
directly reconstruct the sequence $S$. 
However, in practice, we find that this method results in scattered {\it pre-action segments} of short lengths with poor semantic meaning.
Therefore, we propose to achieve this objective through inpainting,
by instantiating an interior decoder $\mathcal{U}$ that predicts the full state sequence from a randomly masked one and the class representation. 
Specifically, we apply a random mask $M \in \mathbb{R}^T$ to the sequence feature $F$ encoded by the linear encoder $E$, which is subsequently concatenated with the discrete latent code $Z$ as the input for the interior decoder, $\mathcal{U}$. 
The reason we use sequence features $F$ instead of the original motion frames $S$ as input is to ensure that gradients can be propagated to the encoder $E$. The output of the interior decoder $\mathcal{U}$ is the completed motion sequence $S'=\{s'_i\}_{i=1}^{T}$. We apply the L2-norm to optimize this discrepancy,
\begin{equation}
\label{eq:in}
    \mathcal{L}^{in} = \left\|\mathcal{U}(F \circ M, Z) - S\right\|_2^2,
\end{equation}
where $\circ$ indicates the masking operation.

\textbf{Boundary Decoding}. 
To further improve the semantic meaningfulness of the discovered action classes, we ask that the discrete action representations be informative about the ending states when an action is completed.
Also, given a segment $p_m$, its final frame $s_{t_m^e}$ is indicative of human switching from one action to another.
Thus, an action representation can be of more semantic significance if it can help distinguish a boundary (action-switching) state from others.  

To achieve this goal, we initialize a boundary decoder, $\mathcal{B}$, to predict the boundary states. 
Similar to the interior decoder, $\mathcal{U}$, the boundary decoder, $\mathcal{B}$, takes the sequence feature, $F$, and the discrete latent code, $Z$, as input, and predict the end-of-segment frame poses, $S^e = \{s_{P(i)}\}_{i=1}^{T}$, where $P(i)$ is the boundary frame timestamp $t^e_m$ of predicted {\it pre-action segment} $p_m$ that contains frame $s_i$. Note that $S^e \in \mathbb{R}^{T\times J}$ has the same dimension as $S$, which are constructed by repeating the predicted boundary poses from the input sequence $S$, i.e., all frames in an action segment are populated with the boundary one. 
And we also apply the L2-norm to optimize the boundary decoder:
\begin{equation}
\label{eq:bound}
    \mathcal{L}^{bound} = \left\|\mathcal{B}(F, Z) - S^{e}\right\|_2^2.
\end{equation}

The {\bf total training loss} $\mathcal{L}^{total}$ for our unsupervised temporal action location pre-training is summarized as:
\begin{equation}
\label{eq:total}
    \mathcal{L}^{total}=\mathcal{L}^{in} + \lambda^{bound} \mathcal{L}^{bound} + \lambda^{com} \mathcal{L}^{com},
\end{equation}
where $\lambda^{bound}$ and $\lambda^{com}$ are predefined weights and studied in the experiments. 

%% file: sections/04_exp.tex
\section{Experiments}
\label{sec:exps}
\subsection{Experiment Setup}
To evaluate our proposed method, we pre-train and transfer our model as follows: We first pre-train the whole network on a large-scale untrimmed skeleton-based dataset without annotation. Then, we finetune the network on an annotated dataset by adding a linear classifier (a residual block \cite{he2016deep} followed by a softmax layer) after the pre-trained linear encoder $E$ without the decoders $\mathcal{U}$ and $\mathcal{B}$. 

\textbf{Datasets}. 
We verify our method on the benchmark dataset, BABEL \cite{punnakkal2021babel}. Most of the temporal action localization datasets are vision-based. To the best of our knowledge, BABEL is the only 3D human motion dataset with frame-level action annotation, which contains about 43 hours of motion sequences and over 250 action categories from AMASS \cite{mahmood2019amass}. We follow S-WTAL \cite{yu2023frame} that create 3 subsets of BABEL, each of which consists of 4 action categories. We pre-train our model on all of the 3 subsets and apply fine-tuning for the subsets separately. 

\textbf{Implementation details}. 
We utilize a Temporal Convolutional Network (TCN) structure for both the encoder and decoders. Our model is trained using the Adam optimizer with an initial learning rate of $10^{-3}$. This rate undergoes a warm-up over the first 20 epochs and is then reduced to one-tenth of its value at the 20th and the 40th epochs. The training spans 500 epochs for both pre-training and fine-tuning phases. We utilize a batch size of 128, and the training is executed on a single NVIDIA Tesla A100 GPU.
\chengcheng{What's the difference between iteration and epoch here? Are they the same? If that's the case, how are we using the learning rate from the beginning? If they are not the same, how many total iterations are used, or what's the stopping criteria?}
We set $T$, $J$, $N$, $\lambda^{bound}$ and $\lambda^{com}$ as 120, 75, 64, 1.0, and 0.05, respectively, by default.

\textbf{Comparison baselines}.
We compare our method with S-WTAL \cite{yu2023frame}, a weakly-supervised skeleton-based temporal activity localization method.
Recognizing the predominance of vision-based methods in this field, we adapt vision-based methods CoLA \cite{zhang2021cola}, FAC-Net \cite{huang2021foreground}, and LART \cite{rajasegaran2023benefits} to skeleton-based settings for comparison. CoLA and FAC-Net are both weakly-supervised vision-based TAL algorithms, and LART is the SOTA fully-supervised vision-based TAL method.
We also adapt UP-TAL \cite{zhang2022unsupervised} which is a vision-based unsupervised approach for temporal action localization pre-training. 

\textbf{Evaluation metrics}.
We follow the standard evaluation protocol. We calculate the Mean Average Precisions (mAPs) under different temporal Intersection of Union (IoU) thresholds, by segmenting sequences based on the frame-level prediction following FAC-Net \cite{huang2021foreground}.

\begin{table}[t]
\small
\centering
\caption{Temporal action localization performance comparisons over BABEL. The column Avg indicates the average mAP at IoU thresholds from 0.1 to 0.5.}
\label{tab:iou}
\begin{tabular}{ccccccc}
\hline
\multicolumn{7}{c}{Subset-1} \\ \hline
\multicolumn{1}{c|}{\multirow{2}{*}{Method}} & \multicolumn{6}{c}{Detection mAP @ IoU (\%)} \\
\multicolumn{1}{c|}{} & 0.1 & 0.2 & 0.3 & 0.4 & \multicolumn{1}{c|}{0.5} & Avg \\ \hline
\multicolumn{1}{c|}{CoLA} & 27.40 & 14.63 & 6.43 & 4.03 & \multicolumn{1}{c|}{2.15} & 10.93 \\
\multicolumn{1}{c|}{FAC-Net} & 29.65 & 17.65 & 8.48 & 3.88 & \multicolumn{1}{c|}{2.62} & 12.46 \\
\multicolumn{1}{c|}{LART} & 52.34 & 45.81 & 38.62 & 30.15 & \multicolumn{1}{c|}{23.82} & 38.15 \\
\multicolumn{1}{c|}{S-WTAL} & 48.74 & 39.82 & 33.15 & 27.39 & \multicolumn{1}{c|}{21.70} & 34.16 \\
\multicolumn{1}{c|}{UP-TAL} & 54.83 & 49.72 & 43.70 & 38.05 & \multicolumn{1}{c|}{32.37} & 43.73 \\
\multicolumn{1}{c|}{Ours} & \textbf{59.57} & \textbf{53.47} & \textbf{46.90} & \textbf{41.00} & \multicolumn{1}{c|}{\textbf{35.01}} & \textbf{47.20} \\ \hline
\multicolumn{7}{c}{Subset-2} \\ \hline
\multicolumn{1}{c|}{\multirow{2}{*}{Method}} & \multicolumn{6}{c}{Detection mAP @ IoU (\%)} \\
\multicolumn{1}{c|}{} & 0.1 & 0.2 & 0.3 & 0.4 & \multicolumn{1}{c|}{0.5} & Avg \\ \hline 
\multicolumn{1}{c|}{CoLA} & 40.10 & 26.86 & 19.9. & 14.02 & \multicolumn{1}{c|}{10.32} & 22.24 \\
\multicolumn{1}{c|}{FAC-Net} & 34.18 & 19.84 & 12.95 & 9.03 & \multicolumn{1}{c|}{6.53} & 16.51 \\
\multicolumn{1}{c|}{LART} & 47.74 & 33.47 & 25.90 & 18.98 & \multicolumn{1}{c|}{15.43} & 28.30 \\
\multicolumn{1}{c|}{S-WTAL} & 61.01 & 50.26 & 40.36 & 29.84 & \multicolumn{1}{c|}{19.55} & 40.20 \\
\multicolumn{1}{c|}{UP-TAL} & 62.10 & 57.55 & 49.17 & 41.16 & \multicolumn{1}{c|}{30.49} & 48.09 \\
\multicolumn{1}{c|}{Ours} & \textbf{64.81} & \textbf{61.05} & \textbf{53.39} & \textbf{42.98} &  \multicolumn{1}{c|}{\textbf{35.13}} &  \textbf{51.47}\\ \hline
\multicolumn{7}{c}{Subset-3} \\ \hline
\multicolumn{1}{c|}{\multirow{2}{*}{Method}} & \multicolumn{6}{c}{Detection mAP @ IoU (\%)} \\
\multicolumn{1}{c|}{} & 0.1 & 0.2 & 0.3 & 0.4 & \multicolumn{1}{c|}{0.5} & Avg \\ \hline
\multicolumn{1}{c|}{CoLA} & 21.50 & 17.40 & 15.16 & 12.10 & \multicolumn{1}{c|}{8.99} & 15.03 \\
\multicolumn{1}{c|}{FAC-Net} & 27.51 & 22.26 & 17.64 & 14.05 & \multicolumn{1}{c|}{8.90} & 18.07 \\
\multicolumn{1}{c|}{LART} & 40.67 & 32.25 & 24.56 & 20.03 & \multicolumn{1}{c|}{16.38} & 26.78 \\
\multicolumn{1}{c|}{S-WTAL} & 35.81 & 31.45 & 26.55 & 23.42 & \multicolumn{1}{c|}{20.36} & 27.52 \\
\multicolumn{1}{c|}{UP-TAL} & 40.20 & 37.15 & 31.94 & 26.85 & \multicolumn{1}{c|}{\textbf{25.19}} & 32.27 \\
\multicolumn{1}{c|}{Ours} & \textbf{42.81} & \textbf{37.96} & \textbf{33.54} & \textbf{29.47} &  \multicolumn{1}{c|}{24.60} & \textbf{33.68} \\ \hline
\end{tabular}
\end{table}

\begin{table*}[t]
\centering
\small
\caption{Results of analysis for different action categories.}
\label{tab:action-class}
\begin{tabular}{l|llll|llll|llll}
\hline
\multicolumn{1}{c|}{\multirow{2}{*}{Method}} & \multicolumn{12}{c}{Detection mAP @ IoU (\%)} \\
 & Walk & Stand & Turn & Jump & Sit & Run & Stand up & Kick & Jog & Wave & Dance & Gesture \\ \hline
CoLA & 43.87 & 37.98 & 14.68 & 13.07 & 78.40 & 40.35 & 19.20 & 22.45 & 23.03 & 5.93 & 43.67 & 13.36 \\
FAC-Net & 49.17 & 31.65 & 22.52 & 15.24 & 57.11 & 36.59 & 16.44 & 26.57 & 40.45 & 15.28 & 42.56 & 11.76 \\
LART & 74.57 & 65.78 & \textbf{41.04} & 27.97 & \textbf{85.70} & 33.13 & \underline{43.07} & 29.07 & 52.36 & \underline{25.54} & 61.91 & \textbf{22.94} \\ 
S-WTAL & 79.37 & \textbf{74.32} & 24.28 & 17.03 & \underline{83.74} & \textbf{57.49} & 23.59 & \underline{79.21} & 45.45 & 16.93 & \textbf{66.13} & 14.75 \\
UP-TAL & \underline{79.93} & 70.15 & 33.44 & \underline{35.77} & 81.83 & 51.21 & 46.45 & 68.92 & \underline{59.29} & \textbf{29.04} & 50.06 & 22.42 \\ \hline
Ours & \textbf{81.86} & \underline{71.81} & \underline{38.87} & \textbf{45.71} & 77.10 & \underline{54.86} & \textbf{48.00} & \textbf{79.28} & \textbf{61.52} & 21.20 & \underline{65.91} & \underline{22.79} \\ \hline
\end{tabular}
\end{table*}

\subsection{Quantitative Analysis}
We show the mAP of the detected action segments on the three subsets in Table \ref{tab:iou}. 
Addressing the complexities of skeleton-based temporal action localization, our approach aligns with the benchmark set by S-WTAL \cite{yu2023frame}, evaluating mAP at thresholds ranging from 0.1 to 0.5, along with their average. In our experiments, our method consistently outperforms other baseline approaches. 

CoLA \cite{zhang2021cola} and FAC-Net \cite{huang2021foreground} are designed for weakly-supervised vision-based tasks. Thus, extracting the necessary information from visual elements is a key factor in their design. Simply altering the feature extractor of these methods to fit the skeleton-based task leads to poor performance. 
LART \cite{rajasegaran2023benefits} is a fully supervised method that combines the skeleton and visual information for TAL. The absence of vision-based data also impacts the performance of LART on our test set.
S-WTAL \cite{yu2023frame} utilizes sequence-level labels to learn the TAL method. Considering that the sequence-level features are mainly influenced by the high-frequency and long actions, it encounters serious long-tail effect. Thus, S-WTAL works worse than UP-TAL and our method, which exploits the frame-level feature for pre-training.
Although UP-TAL \cite{zhang2022unsupervised} is also vision-based, it not only focuses on visual context but also emphasizes the continuity of action frames, making it more effective for skeleton-based tasks, thus outshining other vision-based methods.

Moreover, we detail the mAP at a threshold of 0.1 for different action classes in \ref{tab:action-class}. 
Considering the diverse duration and quantities of different action categories in the dataset, it is crucial to address the issue of balancing different action categories during the training process and mitigate the long-tail effect.
S-WTAL, which relies on sequence-level annotation for training, is notably susceptible to this variation in class distribution. 
\chengcheng{I changed from video-level to sequence level; please confirm.}
In contrast, our method of unsupervised pre-training adeptly captures the intrinsic structure of raw data, mitigating the long-tail effect induced by data distribution. 
Further, our approach surpasses UP-TAL because we learn the segmentation and classification based on the original sequence rather than discontinuous pasted sequences. This enables our network to acquire features tailored to action classes with fewer samples while maintaining robust performance across more prevalent categories.
\chengcheng{I changed ``video'' to sequence in the previous sentence; please verify.}

\textbf{Validation for the number of action discrete codes}. 
We further evaluate the influence of the number of {\it pre-action classes} for VQ layer. Too few classes prevent the network from distinguishing different actions during pre-training, but too many codes cause the distinguished actions to be too short, and lacking in semantics. The work related to human motion reconstruction \cite{jiang2023motiongpt,zhang2023t2m} typically uses more than 128 codes, allowing for a detailed description of motion within a few adjacent frames. However, for our task, we are focused on actions within a longer time window, and intuitively, we need fewer code numbers.
\chengcheng{Why too many code cause the distinguished actions to be too short, instead of just becoming more diverse but equally long?}

We test five different settings of pre-action class numbers $K$: 16, 32, 64, 128, and 256 on three subsets of BABLE \cite{punnakkal2021babel} and their union set, as shown in table \ref{tab:codes}.
We observe that for these subsets, choosing $K=64$ is shown to be more effective for most settings. Too much or too little code can lead to a decrease in algorithm performance.

\begin{table}[htb]
\small
\vspace{-0.5cm}
\centering
\caption{Numerical results on the different number of pre-action classes.}
\label{tab:codes}
\begin{tabular}{ccccccc}
\hline
\multicolumn{7}{c}{Subset-1} \\ \hline
\multicolumn{1}{c|}{\multirow{2}{*}{Method}} & \multicolumn{6}{c}{Detection mAP @ IoU (\%)} \\
\multicolumn{1}{c|}{} & 0.1 & 0.2 & 0.3 & 0.4 & \multicolumn{1}{c|}{0.5} & Avg \\ \hline
\multicolumn{1}{c|}{$K=16$} & 56.54 & \underline{53.12} & \textbf{48.17} & \textbf{42.88} & \multicolumn{1}{c|}{\textbf{36.95}} & \textbf{47.53} \\
\multicolumn{1}{c|}{$K=32$} & 56.99 & 51.91 & 46.34 & 40.63 & \multicolumn{1}{c|}{34.05} & 45.99 \\
\multicolumn{1}{c|}{$K=64$} & \textbf{59.57} & \textbf{53.47} & 46.90 & 41.00 & \multicolumn{1}{c|}{\underline{35.01}} & \underline{47.20} \\
\multicolumn{1}{c|}{$K=128$} & \underline{58.36} & 53.09 & \underline{47.60} & \underline{41.48} & \multicolumn{1}{c|}{34.28} & 46.96 \\
\multicolumn{1}{c|}{$K=256$} & 57.10 & 52.24 & 46.80 & 40.71 & \multicolumn{1}{c|}{34.09} & 46.19 \\\hline
\multicolumn{7}{c}{Subset-2} \\ \hline
\multicolumn{1}{c|}{\multirow{2}{*}{Method}} & \multicolumn{6}{c}{Detection mAP @ IoU (\%)} \\
\multicolumn{1}{c|}{} & 0.1 & 0.2 & 0.3 & 0.4 & \multicolumn{1}{c|}{0.5} & Avg \\ \hline
\multicolumn{1}{c|}{$K=16$} & 60.71 & 55.24 & 48.12 & 42.04 & \multicolumn{1}{c|}{33.63} & 47.95 \\
\multicolumn{1}{c|}{$K=32$} & 62.82 & 57.98 & \underline{52.42} & \textbf{43.64} & \multicolumn{1}{c|}{\underline{35.44}} & \underline{50.86} \\
\multicolumn{1}{c|}{$K=64$} & \textbf{64.81} & \textbf{61.05} & \textbf{53.39} & \underline{42.98} &  \multicolumn{1}{c|}{35.13} &  \textbf{51.47} \\
\multicolumn{1}{c|}{$K=128$} & \underline{63.67} & \underline{59.31} & 50.87 & 43.28 & \multicolumn{1}{c|}{\textbf{35.58}} & 50.54 \\
\multicolumn{1}{c|}{$K=256$} & 61.31 & 58.37 & 49.28 & 40.36 & \multicolumn{1}{c|}{32.99} & 48.46 \\ \hline
\multicolumn{7}{c}{Subset-3} \\ \hline
\multicolumn{1}{c|}{\multirow{2}{*}{Method}} & \multicolumn{6}{c}{Detection mAP @ IoU (\%)} \\
\multicolumn{1}{c|}{} & 0.1 & 0.2 & 0.3 & 0.4 & \multicolumn{1}{c|}{0.5} & Avg \\ \hline
\multicolumn{1}{c|}{$K=16$} & 40.02 & 35.49 & 28.80 & 24.32 & \multicolumn{1}{c|}{20.04} & 29.73 \\
\multicolumn{1}{c|}{$K=32$} & \textbf{43.48} & 36.68 & \underline{28.95} & \underline{26.92} & \multicolumn{1}{c|}{\underline{23.71}} & \underline{31.95} \\
\multicolumn{1}{c|}{$K=64$} & \underline{42.81} & \textbf{37.96} & \textbf{33.54} & \textbf{29.47} &  \multicolumn{1}{c|}{\textbf{24.60}} & \textbf{33.68} \\
\multicolumn{1}{c|}{$K=128$} & 37.38 & 32.74 & 26.07 & 22.61 & \multicolumn{1}{c|}{19.31} & 27.62 \\
\multicolumn{1}{c|}{$K=256$} & 41.23 & \underline{37.03} & 28.65 & 25.47 & \multicolumn{1}{c|}{20.96} & 30.67 \\ \hline

\end{tabular}
\end{table}

\textbf{Ablation study}. To further validate the design of our algorithm, we have designed the following ablation study:
\begin{itemize}[leftmargin=*]
    \setlength{\itemsep}{0pt}
    \setlength{\parsep}{0pt}
    \setlength{\parskip}{0pt}

    \item \textbf{Ours - w/o RVQ}: We remove the RVQ layers and utilize $Z^0$ as the output discrete latent code to validate the benefits of Residual VQ-VAE.
    \item \textbf{Ours - w/o $M$}: We remove the mask operation for the input of $\mathcal{U}$. In this way, we modify the sequence inpainting task into a reconstruction task.
    \item \textbf{Ours - w/o $\mathcal{U}$}: We delete the interior decoder $\mathcal{U}$ and loss $\mathcal{L}^{in}$ to validate the benefits of interior constraint. 
    \item \textbf{Ours - w/o $\mathcal{B}$}: We delete the boundary decoder $\mathcal{B}$ and loss $\mathcal{L}^{bound}$ to validate the benefits of boundary constraint. 
\end{itemize}
The results are demonstrated in Table~\ref{tab:ablation}. The complete version of our algorithm outperforms the modified versions across different subsets.
When we modify the sequence inpainting task into a reconstruction task by removing the mask operation for $\mathcal{U}$, the performance decreases on all subsets, and it even is weaker than the version where only $\mathcal{B}$ is retained, which demonstrate that reconstruction-based optimization objectives reduce the network's ability to segment and classify different action categories. This is because the reconstruction process leads to shorter {\it pre-action segments} with less semantic clarity, which is not suitable for TAL task.
Besides, the imbalance of detail information and action semantics and the problem of codebook collapse caused by a single code book, make the network without RVQ layers of the network significantly weaker than other settings in Sets 1 and 3.
Meanwhile, both optimization objectives are beneficial to the results, and the removal of any one of them leads to a decline in the performance of the algorithm.
These experiments indicate that each key component of our algorithm design contributes to its enhancement.

\begin{table}[htb]
\small
\vspace{-0.35cm}
\centering
\caption{Numerical results for ablation study}
\label{tab:ablation}
\begin{tabular}{cccclll}
\hline
\multicolumn{7}{c}{Average detection mAP @ IoU (\%)} \\ \hline
\multicolumn{1}{c|}{\multirow{2}{*}{Method}} & \multicolumn{6}{c}{Detection mAP @ IoU (\%)} \\
\multicolumn{1}{c|}{} &  &  & \multicolumn{4}{c}{} \\ \hline
\multicolumn{1}{c|}{Ours - recon} & 46.63 & 48.74 & \multicolumn{4}{c}{31.22} \\
\multicolumn{1}{c|}{Ours - w/o RVQ} & 45.29 & 50.55 & \multicolumn{4}{c}{27.58} \\
\multicolumn{1}{c|}{Ours - w/o $\mathcal{U}$} & 46.82 & 48.72 & \multicolumn{4}{c}{31.43} \\
\multicolumn{1}{c|}{Ours - w/o $\mathcal{B}$} & 46.97 & 49.15 & \multicolumn{4}{c}{31.19} \\ \hline
\multicolumn{1}{c|}{Ours} & \textbf{47.20} & \textbf{51.47} & \multicolumn{4}{c}{\textbf{33.68}} \\ \hline
\end{tabular}
\end{table}

\begin{figure}[htb]
    \centering
    \includegraphics[width=\linewidth]{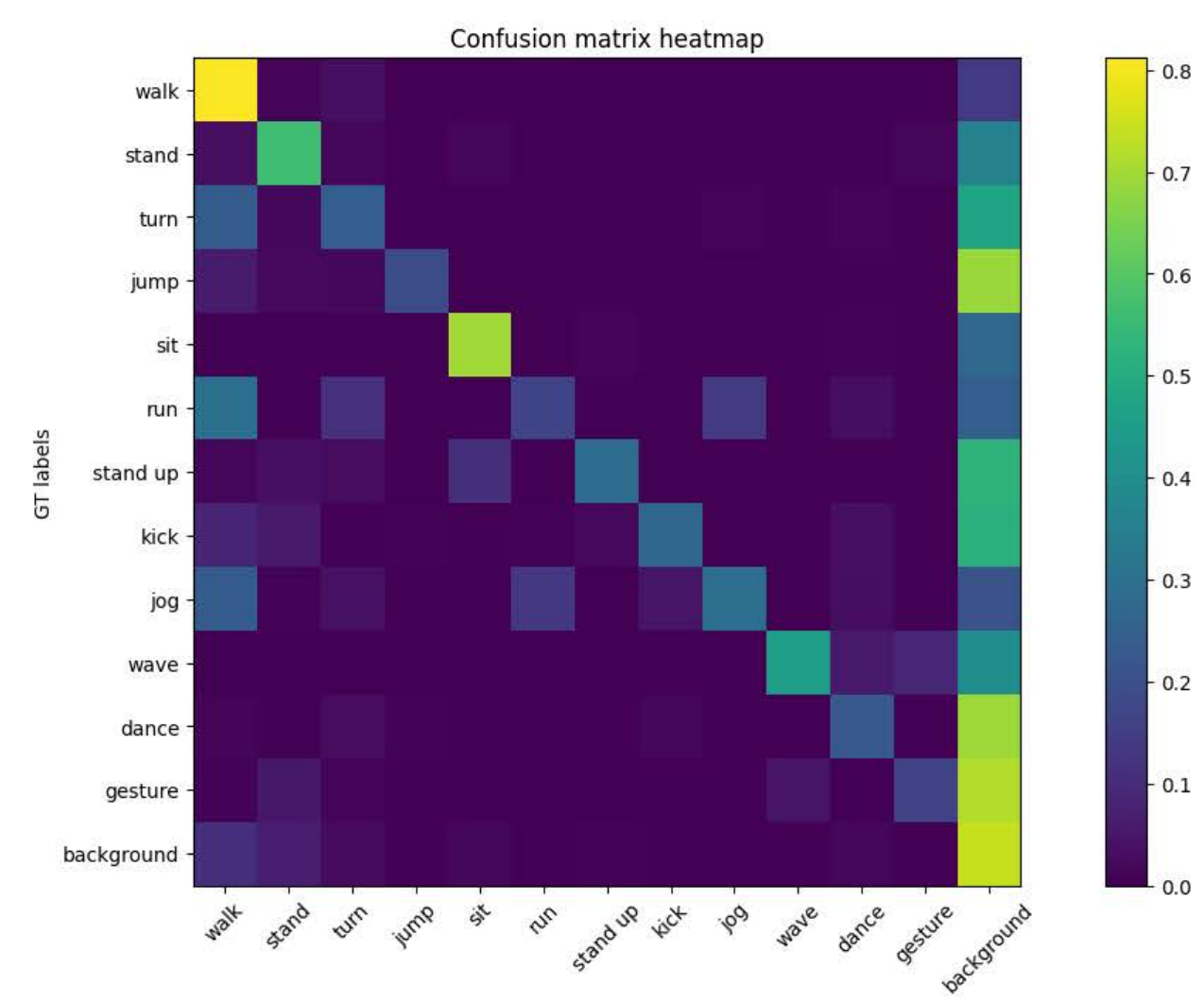}
    \caption{We calculate the confusion matrix comparing the predicted action classes to the ground truth to identify the bottlenecks of our method. We conduct evaluation on the union set of the three subsets, including 12 action classes.}
    \label{fig:confusion}
\end{figure}

\textbf{Confusion matrix for our prediction and ground truth action classes}.
To further identify the current bottlenecks of our algorithm, we conduct evaluation on the union set of these three subsets, including 12 action classes. Additionally, we compute a confusion matrix comparing the predicted action classes to the ground truth as shown in Figure \ref{fig:confusion} which reveals that the primary factor impacting our algorithm's performance is the background, apart from the targeted 12 action categories. The impact of background is twofold. Firstly, the background occupies a significant portion of the entire sequence and contains a high level of diversity, making its segmentation more challenging compared to other categories. Secondly, different action classes may exhibit similar motion frames. For instance, in BABEL, the classes ``step'' and ``move back to original position'' visually resemble ``walk''. However, ``step'' and ``move back to original position'' are listed in the background, which introduces label noises for background segmentation.

%% file: sections/05_conclusion.tex
\section{Conclusion}
\label{sec:conclusion}

We propose a novel unsupervised pre-training method for skeleton-based temporal action localization. 
We consider different actions as different flow fields and represent these flow fields as discrete latent codes. To make these flow fields semantically meaningful, we design two optimization objectives to encourage these discrete latent codes to be informative about the state transition within an action and the ending state of that action. Further, we demonstrate the benefits of the latent space learned through our method for temporal action localization, both through comparison with the baselines and through ablation experiments.

 
However, our approach is still limited in some aspects. 
\begin{itemize}[leftmargin=*]
    \setlength{\itemsep}{0pt}
    \setlength{\parsep}{0pt}
    \setlength{\parskip}{0pt}
    \item When the motion data exhibits significant diversity in the background, our algorithm encounters difficulties in establishing a clear and distinct decision boundary for background differentiation.
    \item Further, as we focus on skeleton-based temporal action localization, extending the application of the action flow field to vision-based or hybrid tasks remains a substantial challenge.
    \item Finally, the scarcity of action-specific skeleton-based datasets limits broader applications of our framework.

\end{itemize}

%% file: sections/06_appendix.tex
\newpage
\appendix
\onecolumn

\section{Implementation Details}
We apply a two-codebook Residual VQ-VAE structure for the segmentation and {\it pre-action} segmentation. The first codebook $c^{VQ}$ is set for VQ layer, while its quantization results are considered as the {\it pre-action class}. The second codebook $c^{RVQ}$ is shared by all the RVQ layers, which is only used for the action flow field decoders $\mathcal{U}$ and $\mathcal{B}$. In most of our experiments, we utilize 4 RVQ layers, and the code numbers of $c^{VQ}$ and $c^{RVQ}$ are 64 and 256, respectively. 
The codebooks are updated by exponential moving averages (EMA).

Besides, to improve the usage of these codebooks, and avoid the discrete latent code $Z$ is the same as continuous feature $F$, we reduce the dimensionality of $F$ to 16 through a linear layer before performing quantization. After quantization, we will use another linear layer to increase its dimensionality back to 256. We calculate the commitment loss based on the features after dimensionality reduction.

We present the pseudo code of the training process of BID in Algorithm~\ref{algo:ours}.

\section{More quantitative analysis}
\begin{table}[htb]
\centering
\caption{Numerical results on the different number of pre-action classes.}
\label{tab:codes}
\begin{tabular}{ccccccc}
\hline
\multicolumn{7}{c}{Subset-Union} \\ \hline
\multicolumn{1}{c|}{\multirow{2}{*}{Method}} & \multicolumn{6}{c}{Detection mAP @ IoU (\%)} \\
\multicolumn{1}{c|}{} & 0.1 & 0.2 & 0.3 & 0.4 & \multicolumn{1}{c|}{0.5} & Avg \\ \hline
\multicolumn{1}{c|}{$K=16$} & 41.80 & \textbf{38.52} & 34.17 & 28.89 & \multicolumn{1}{c|}{23.12} & 33.30 \\
\multicolumn{1}{c|}{$K=32$} & \textbf{42.31} & 38.14 & \underline{34.28} & \underline{29.23} & \multicolumn{1}{c|}{23.00} & \underline{33.39} \\
\multicolumn{1}{c|}{$K=64$} & \underline{42.09} & \underline{38.23} & \textbf{34.58} & \textbf{30.27} &  \multicolumn{1}{c|}{\textbf{24.15}} & \textbf{33.86} \\
\multicolumn{1}{c|}{$K=128$} & 41.57 & 37.08 & 33.33 & 28.74 & \multicolumn{1}{c|}{\underline{23.01}} & 32.74 \\
\multicolumn{1}{c|}{$K=256$} & 39.81 & 35.62 & 32.15 & 27.36 & \multicolumn{1}{c|}{22.20} & 31.43 \\ \hline
\end{tabular}
\end{table}

We evaluate the influence of the number of {\it pre-action classes} in the VQ layers on the union subset of the three subsets we used in Section 5 with 12 action classes. An interesting observation is that despite the union set requiring the detection of 12 actions, as opposed to just 4 in individual subsets, our choice of $K$ remained consistent in yielding similar conclusions. $K=64$ is still the best choice for the experiments.
The reason is that an optimal choice of $K$ enables better-distinguishing action segments during pre-training, a distinction that universally benefits downstream tasks.

\begin{figure}[H]
    \centering
    \includegraphics[width=0.5\linewidth]{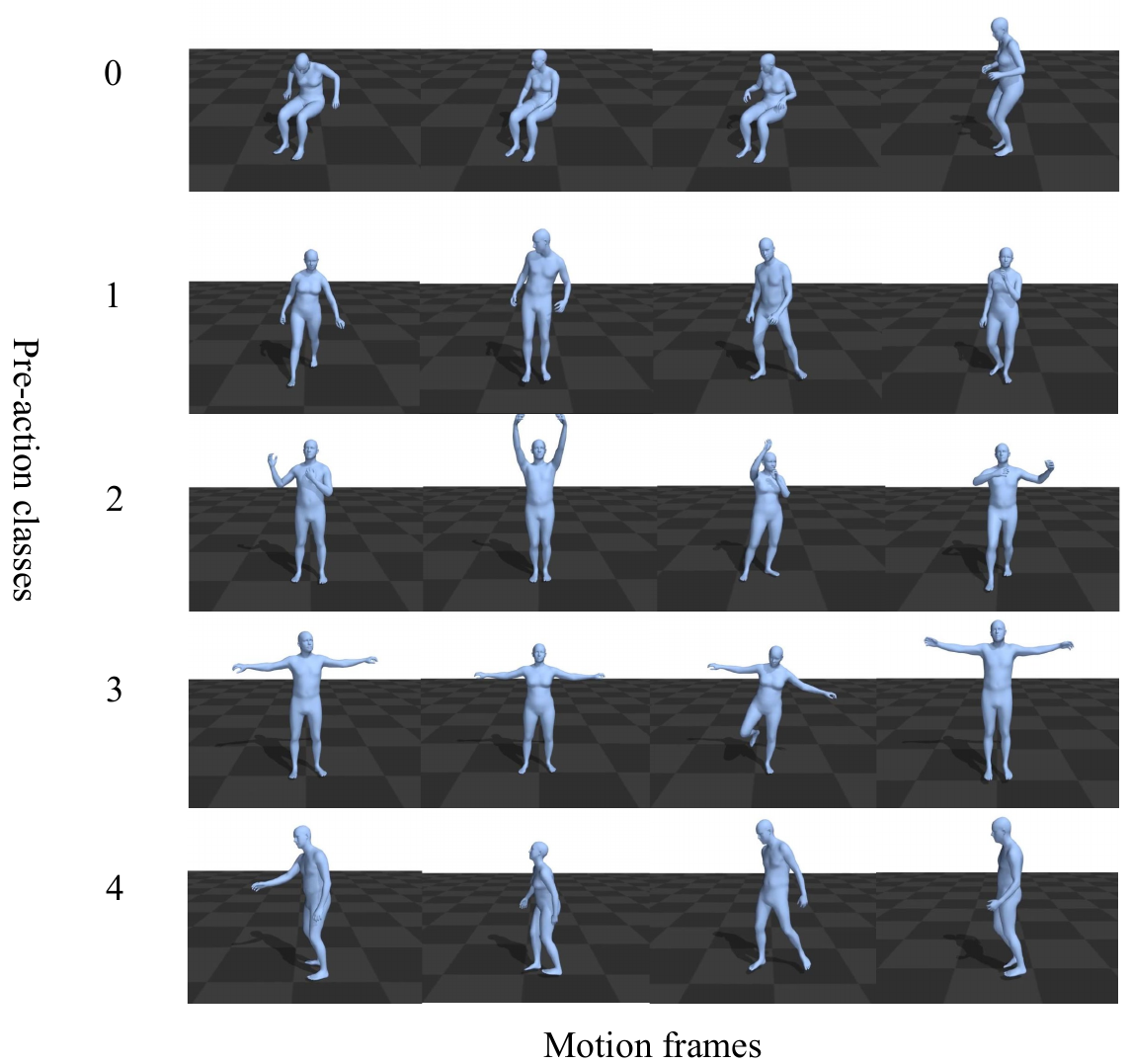}
    \caption{Visualization for pre-action classes and their corresponding motion frames.}
    \label{fig:cluster}
\end{figure}

\section{Visual results}
\subsection{Visualization for Pre-action Classes}

In Figure~\ref{fig:cluster}, we present some {\it pre-action classes} learned by our unsupervised pre-training method with their corresponding motion frames. We can observe that these {\it pre-action classes} are related to specific human body movements, such as sitting, arm lifting, and T-pose. Through these {\it pre-action classes}, our pre-training model can segment input motion sequences into meaningful segments. Through the fine-tuning process, this ability of segmentation and classification are transferred to the domain of human annotation.

\subsection{Visual Results for Pre-Training}

\begin{figure*}[htbp]
    \centering
    \includegraphics[width=1\linewidth]{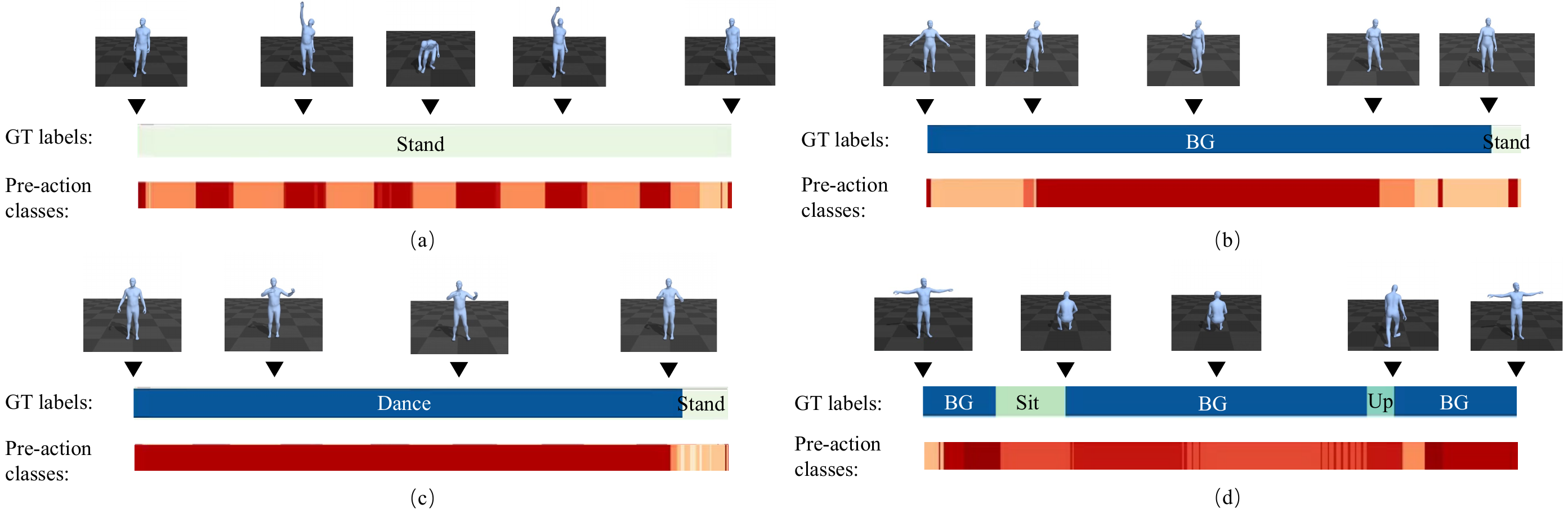}
    \caption{Visualization for discrete encode results of pre-training model}
    \label{fig:pre-train}
\end{figure*}

In Figure~\ref{fig:pre-train}, we compare some {\it pre-action segmentation} predicted by our unsupervised pre-training model with the corresponding ground-truth segment, where different colors represent different {\it pre-action classes} or ground-truth labels. We observe that, despite the absence of ground truth supervision during the pre-training process, our model still achieves segmentation results quite close to the ground truth. Furthermore, our pre-training process is capable of learning action patterns not annotated in the ground truth labels, such as the repetitive bending and raising of hands by the character in sequence (a). Besides, we include a video for visualizing the pre-training results, further demonstrating the information learned by our algorithm.    

\begin{figure}[htb]
    \centering
    \includegraphics[width=0.68\linewidth]{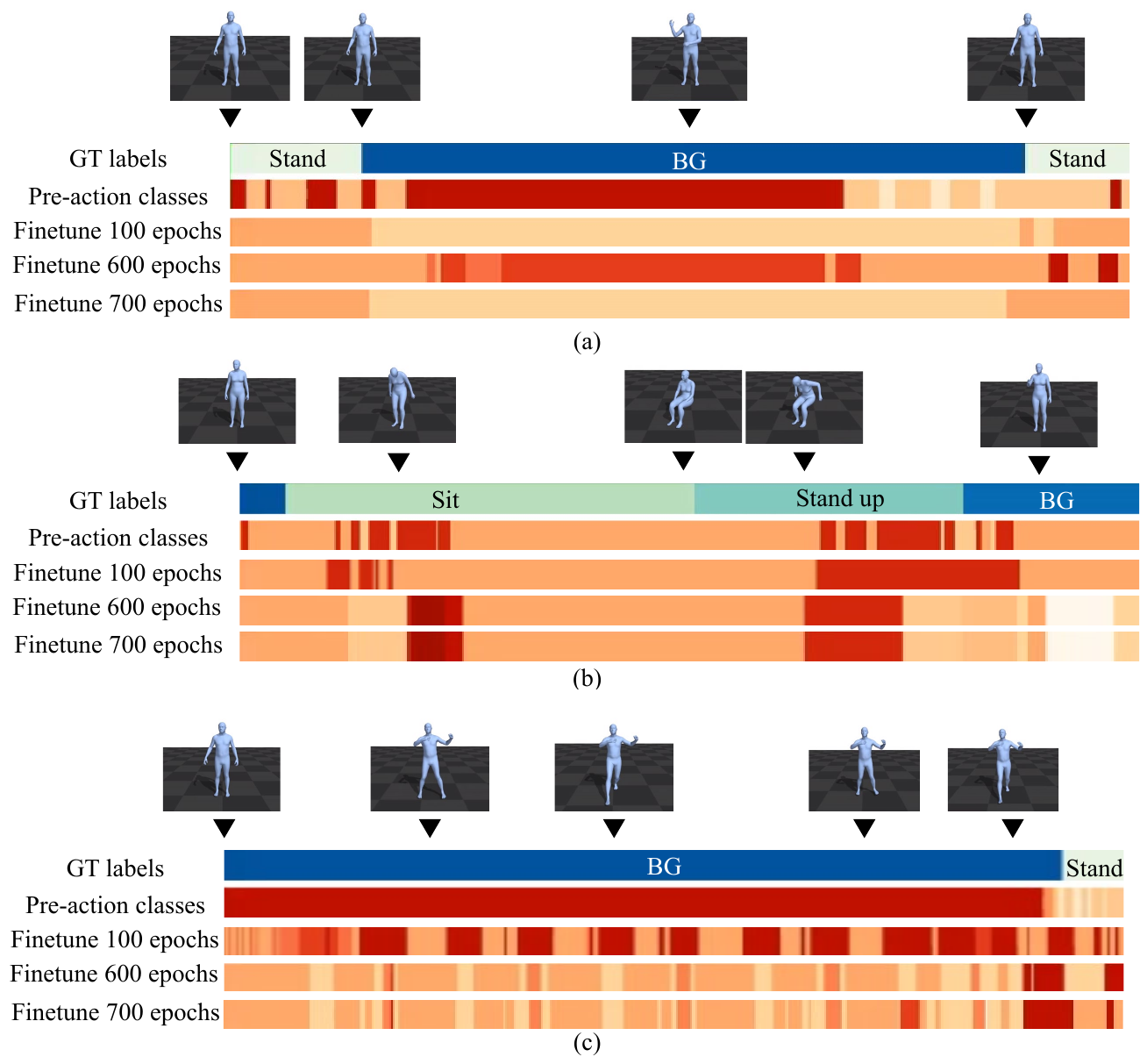}
    \caption{Visualization of discrete encode results for finetuning model}
    \label{fig:finetune}
\end{figure}

\subsection{Visualization for Fine-Tuning Process}

To enable fine-tuning process, we add a linear classifier after the encoder $E$. We want to understand the changes in features during the fine-tuning process. Thus, we utilize freeze pre-trained VQ layer and codebook $c^{VQ}$ to quantize the features outputted during the fine-tuning process.

We present some samples in Figure~\ref{fig:finetune}. In our initial guess, the finetuning process mainly involved the update of features and the aggregation of adjacent segments. 
Sequence (a) aligns with our original guess. However, we found that this pattern does not apply to all sequences. 
For instance, in sequence (b), the pre-trained model attempts to predict two rough transition processes (red segments) for sit state. As fine-tuning progresses, these transitions are not absorbed into the 'sit' or 'stand up' but become more precise, even though the ground truth does not provide information about the transitions. 
In sequence (c), we further discovered that even though our network could roughly segregate the background of an action during pre-training, it is still capable of further dissecting the background through the finetuning process. 
We believe this indicates that our pre-training process has learned a representation that can effectively express different action classes, which can be refined through a labeled fine-tuning process, resulting in accurate and fine-grained action representations.

\section{Statistical analysis}
\begin{figure}[H]
    \centering
    \includegraphics[width=0.3\linewidth]{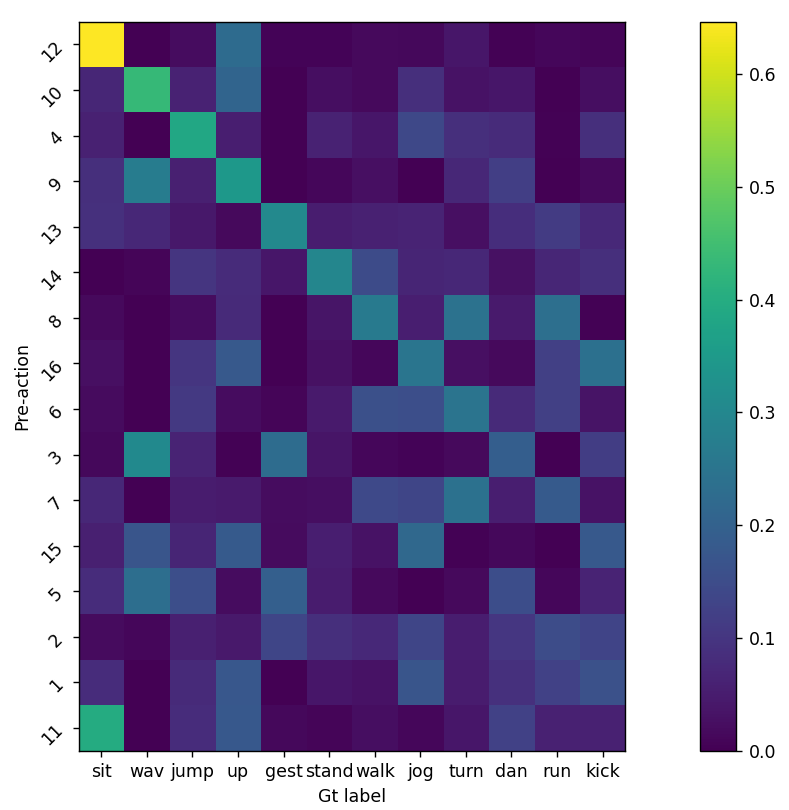}
    \caption{Confusion map for pre-action classes and ground truth action classes.}
    \label{fig:cmap}
\end{figure}

We learn a discrete encoder to segment the input motion sequences and classify the segments as pre-action classes. Although these pre-action classes are not equivalent to annotated action classes, we still hope there is a correlation between them. We present the confusion map for pre-action classes and ground truth labels, where we set $K=16$, in figure \ref{fig:cmap}. We found that some pre-action classes have a stronger association with a few specific action labels, like pre-action 12 is highly related to action sit and stand up. Some pre-action classes have connections with multiple different actions, like pre-actions 1 and 2.

\begin{algorithm}
\caption{The training process of BID}\label{alg:cap}
\label{algo:ours}
\begin{algorithmic}[1]
    \STATE \textbf{Input:} Skeleton-based motion sequence $S={s_i} \in \mathbb{R}^{T \times D}$, boundary loss weight $\lambda^{bound}$ and commitment loss weight $\lambda^{com}$.
    \STATE \textbf{Initialization:} Linear encoder $E$,  VQ layer $VQ$, RVQ layers ${RVQ^l}_{l=0}^L$, Interior decoder $\mathcal{U}$ and Boundary decoder $\mathcal{B}$.    
    \STATE $F \leftarrow E(S)$  (Equation~\ref{eq:encoding}) 
    \STATE $Z^0 \leftarrow VQ(F)$
    \STATE Segment $S$ into $\{p_m\}_{m=0}^{M}$ based on $Z^0$
    \STATE Copy end frames of $\{p_m\}_{m=0}^{M}$ to get $S^e$
    \STATE $l \leftarrow 1$

    \WHILE{$l \leq L$}
        \STATE $r^l \leftarrow F-Z^0-\sum\limits_{j=0}^{L-1}R^j$
        \STATE $R^l \leftarrow RVQ^{l}(r^l)$
    \ENDWHILE

    \STATE $Z = Z^0 + \sum\limits_{j=1}^{L}R^j$ (Equation~\ref{eq:sumz}) 
    \STATE $\mathcal{L}^{com} \leftarrow \left\|F-Z\right\|^2_2$ (Equation~\ref{eq:com}) 
    \STATE Apply EMA update to the codebook of $VQ$ and $RVQ$

    \STATE Generate random mask $M \in \mathbb{R}^T$
    \STATE $S' \leftarrow \mathcal{U}(F \circ M, Z)$
    \STATE $\mathcal{L}^{in} \leftarrow \left\|S-S'\right\|^2_2$ (Equation~\ref{eq:in}) 
    
    \STATE $S^{e'} \leftarrow \mathcal{B}(F, Z)$
    \STATE $\mathcal{L}^{bound} \leftarrow \left\|S^e-S^{e'}\right\|^2_2$ (Equation~\ref{eq:bound}) 

    \STATE $L^{total} \leftarrow \mathcal{L}^{in} + \lambda^{bound} \mathcal{L}^{bound} + \lambda^{com} \mathcal{L}^{com}$ (Equation~\ref{eq:total}) 
        
\end{algorithmic}
\end{algorithm}

\section{Architecture of Our Network}
We utilize TCN-based structures for both encoder and decoders, which extracts spatio features for human motion through convolution with a kernel size of 1, and extract temporal features through dilation convolution and larger kernels (9 or 3).
We list the details of our network architecture in Table~\ref{tab:architecture}. The decoders $\mathcal{U}$ and $\mathcal{B}$ share the same architecture.

\section{Symbol meaning}
We list the symbols we used in the paper and their meaning in Table~\ref{tab:symbols}.

\begin{figure}[htb]
    \centering
    \includegraphics[width=0.6\linewidth]{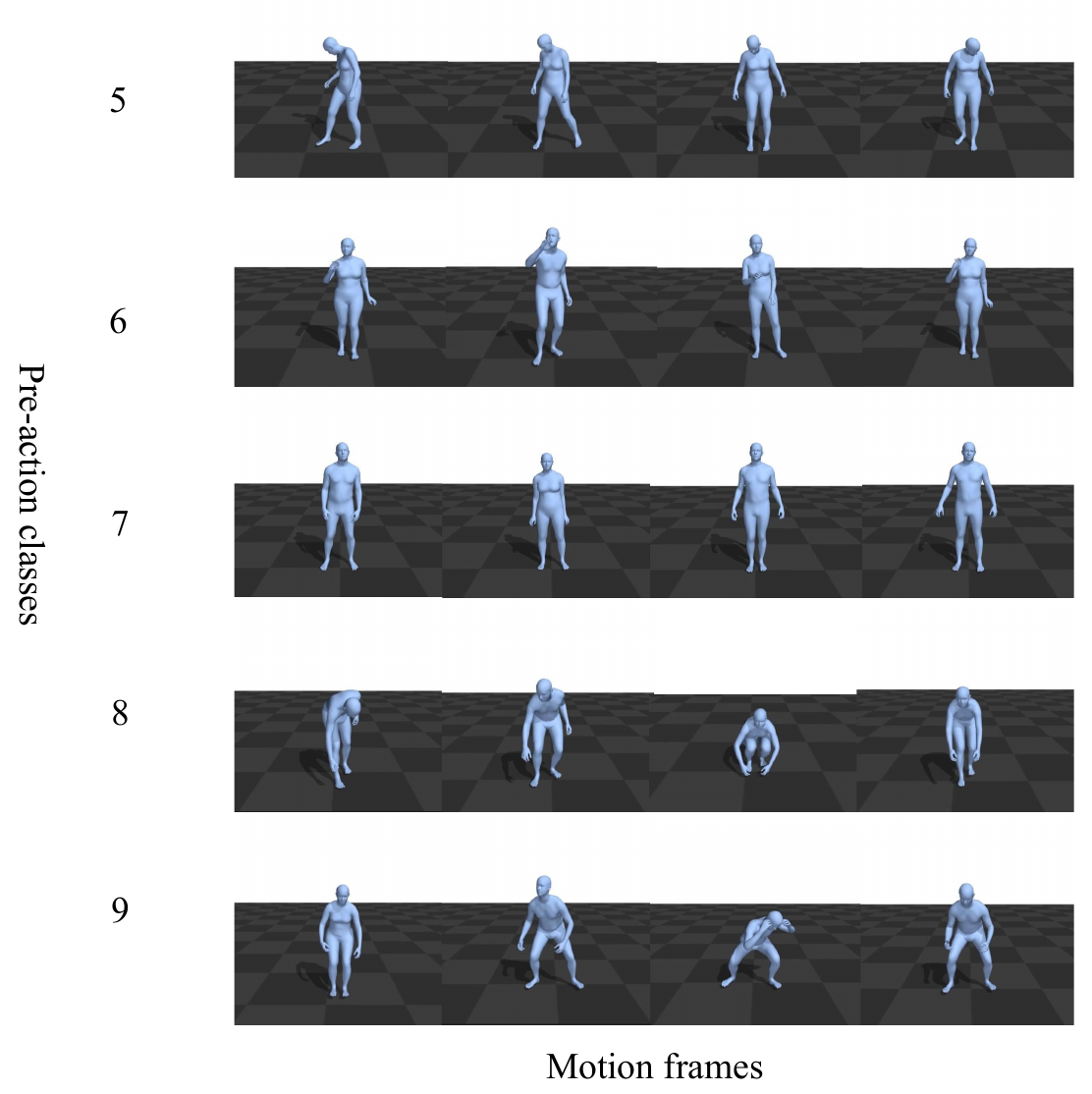}
    \caption{More visualization for pre-action classes and their corresponding motion frames.}
    \label{fig:cluster2}
\end{figure}
\section{More visualization}
We show more visualization in Figure~\ref{fig:cluster2}

\begin{table}[htb]
\centering
\caption{Architecture of our Method}
\label{tab:architecture}
\begin{tabular}{ll}
\hline
Components & Architecture \\ \hline
Linear Encoder & \begin{tabular}[c]{@{}l@{}}
(0): Conv1D(J*3, 256, kernel\_size=(3,), stride=(1,), padding=(1,))\\ 
(1): ReLU()\\ 
(2): 2 $\times$ Sequential(\\   
\ \ \ \ (0): Conv1d(256, 256, kernel\_size=(3,), stride=(1,), padding=(1,))\\   
\ \ \ \ (1): ResConv1DBlock(\\     
\ \ \ \ \ \ \ \ (0): (activation1): ReLU()\\     
\ \ \ \ \ \ \ \ (1): (conv1): Conv1D(256, 256, kernel\_size=(3,), stride=(1,), padding=(9,), dilation=(9,))\\     
\ \ \ \ \ \ \ \ (2): (activation2): ReLU()\\     
\ \ \ \ \ \ \ \ (3): (conv2): Conv1D(256, 256, kernel\_size=(1,), stride=(1,)))\\     
\ \ \ \ (2): ResConv1DBlock(\\         
\ \ \ \ \ \ \ \ (0): (activation1): ReLU()\\         
\ \ \ \ \ \ \ \ (1): (conv1): Conv1D(256, 256, kernel\_size=(3,), stride=(1,), padding=(3,), dilation=(3,))\\         
\ \ \ \ \ \ \ \ (2): (activation2): ReLU()\\         
\ \ \ \ \ \ \ \ (3): (conv2): Conv1D(256, 256, kernel\_size=(1,), stride=(1,)))\\     
\ \ \ \ (3): ResConv1DBlock(\\         
\ \ \ \ \ \ \ \ (0): (activation1): ReLU()\\         
\ \ \ \ \ \ \ \ (1): (conv1): Conv1D(256, 256, kernel\_size=(3,), stride=(1,), padding=(1,))\\         
\ \ \ \ \ \ \ \ (2): (activation2): ReLU()\\         
\ \ \ \ \ \ \ \ (3): (conv2): Conv1D(256, 256, kernel\_size=(1,), stride=(1,))))\end{tabular} \\ \hline
Residual VQ & \begin{tabular}[c]{@{}l@{}}
(0): (conv1): Conv1D(256, 16, kernel\_size=(1,), stride(1,))\\ 
(1): (codebook\_class): nn.Parameter((64, 16), requires\_grad=False)\\ 
(2): (codebook\_residual): nn.Parameter((64, 16), requires\_grad=False)\\ (3): (conv2): Conv1d: Conv1D(16, 256, kernel\_size=(1,), stride=(1,))\end{tabular} \\ \hline
Decoder & \begin{tabular}[c]{@{}l@{}}
(0): 2 $\times$ Sequential(\\   
\ \ \ \ (0): Conv1d(256, 256, kernel\_size=(3,), stride=(1,), padding=(1,))\\   
\ \ \ \ (1): ResConv1DBlock(\\     
\ \ \ \ \ \ \ \ (0): (activation1): ReLU()\\     
\ \ \ \ \ \ \ \ (1): (conv1): Conv1D(256, 256, kernel\_size=(3,), stride=(1,), padding=(9,), dilation=(9,))\\     
\ \ \ \ \ \ \ \ (2): (activation2): ReLU()\\     
\ \ \ \ \ \ \ \ (3): (conv2): Conv1D(256, 256, kernel\_size=(1,), stride=(1,)))\\     
\ \ \ \ (2): ResConv1DBlock(\\         
\ \ \ \ \ \ \ \ (0): (activation1): ReLU()\\         
\ \ \ \ \ \ \ \ (1): (conv1): Conv1D(256, 256, kernel\_size=(3,), stride=(1,), padding=(3,), dilation=(3,))\\         
\ \ \ \ \ \ \ \ (2): (activation2): ReLU()\\         
\ \ \ \ \ \ \ \ (3): (conv2): Conv1D(256, 256, kernel\_size=(1,), stride=(1,)))\\     
\ \ \ \ (3): ResConv1DBlock(\\         
\ \ \ \ \ \ \ \ (0): (activation1): ReLU()\\         
\ \ \ \ \ \ \ \ (1): (conv1): Conv1D(256, 256, kernel\_size=(3,), stride=(1,), padding=(1,))\\         
\ \ \ \ \ \ \ \ (2): (activation2): ReLU()\\         
\ \ \ \ \ \ \ \ (3): (conv2): Conv1D(256, 256, kernel\_size=(1,), stride=(1,)))) \\
\ \ \ \ (2) Conv1D(256, 256, kerne\_size=(1,), stride=(1,)) \\
(1): ReLU() \\
(2): Conv1D(256, 75, kernel\_size=(1,), stride=(1,)) \\
\end{tabular} \\ \hline
\end{tabular}
\end{table}

\begin{table}[htb]
\centering
\caption{The meaning of symbols.}
\label{tab:symbols}
\begin{tabular}{l|l}
\hline
\textbf{Symbol} & \textbf{Meaning} \\ \hline
$\mathcal{S}$ & Skeleton-based motion sequence dataset \\ \hline
$S$ & Skeleton-based motion sequence \\ \hline
$s_i$ & $i$-th frame in sequence $S$ \\ \hline
$T$ & Number of frames in sequence $S$ \\ \hline
$J$ & Number of freedom of skeleton joints \\ \hline
$N$ & Pre-defined number of pre-action class \\ \hline
$M$ & Number of \{pre-action\} segments in sequence $S$ \\ \hline
$p$ & \textit{Pre-action} segment \\ \hline
$p^m$ & $m$-th \textit{pre-action segment} in sequence  $S$ \\ \hline
$t^b_m$ & The begging timestamp of $p_m$ \\ \hline
$t^e_m$ & The ending timestamp of $p_m$ \\ \hline
$a_m$ & The \textit{pre-action class} of $p_m$ \\ \hline
$\mathcal{G}$ & Flow field function \\ \hline
$t$ & Timestamp \\ \hline
$x_t$ & object's position at time $t$. \\ \hline
$\mathcal{G}^{a_m}$ & Flow field function related to {\it pre-action class} $a_m$ \\ \hline
$\mathcal{E}$ & Motion encoding function \\ \hline
$E$ & Linear motion encoder \\ \hline
$F$ & Sequence feature (the output of $E$) \\ \hline
$f_i$ & $i$-th pre-frame feature of $F_i$ (the continuous feature of $s_i$) \\ \hline
$c$ & Codebook \\ \hline
$Z$ & Discrete latent code projected from $F$ \\ \hline
$VQ$ & VQ layer \\ \hline
$c_j$ & j-th discrete latent in $c$ \\ \hline
$sg$ & Stop gradient \\ \hline
$\mathcal{L}^{com}$ & Commitment loss \\ \hline
$RVQ$ & Residual VQ layer \\ \hline
$Z^0$ & Discrete latent code projected from $F$ by VQ layer \\ \hline
$L$ & Number of RVQ layers \\ \hline
$R^l$ & Residual discrete latent code projected from $F - Z^0 - \sum\limits_{j=1}^{j<i-1}{R^j}$ by $l$-th RVQ layer \\ \hline
$c^{VQ}$ & Codebook of VQ layer \\ \hline
$c^{RVQ}$ & Shared codebook of RVQ layers \\ \hline
$M$ & Random mask \\ \hline
$\mathcal{U}$ & Interior decoder \\ \hline
$\mathcal{L}^{in}$ & Interior loss \\ \hline
$\mathcal{B}$ & Boundary decoder \\ \hline
$P(i)$ & The boundary frame timestamp $t^e_m$ of predicted \textit{pre-action segment} $p_m$ tht contains frame $s_i$ \\ \hline
$S^{e}$ & End-of-segment sequence $\{s_{P(i)}\}_{i=1}^{T}$ \\ \hline
$\mathcal{L}^{bound}$ & Boundary loss \\ \hline
$\mathcal{L}^{total}$ & Total loss \\ \hline
$\lambda^{bound}$ & Loss weight for $\mathcal{L}^{bound}$ \\ \hline
$\lambda^{com}$ & Loss weight for $\mathcal{L}^{com}$ \\ \hline
\end{tabular}
\end{table}